\documentclass{article}

\usepackage{arxiv}

\usepackage[utf8]{inputenc} 
\usepackage[T1]{fontenc}    
\usepackage{hyperref}       
\usepackage{url}            
\usepackage{booktabs}       
\usepackage{amsfonts}       
\usepackage{nicefrac}       
\usepackage{microtype}      
\usepackage{graphicx}
\usepackage{natbib}
\usepackage{doi}

\usepackage{epsfig}
\usepackage{subcaption}
\usepackage{calc}
\usepackage{amssymb}
\usepackage{amstext}
\usepackage{amsmath}
\usepackage{amsthm}
\usepackage{multicol}
\usepackage{pslatex}
\usepackage{apalike}
\usepackage[bottom]{footmisc}
\usepackage{adjustbox}
\usepackage{pgf}
\usepackage{float}
\floatstyle{plaintop}
\restylefloat{table}
\usepackage[acronym, toc]{glossaries}

\title{How far generated data can impact Neural Networks performance?}

\date{February, 2023} 					
\author{ \href{https://orcid.org/0000-0003-2675-5463}{\includegraphics[scale=0.06]{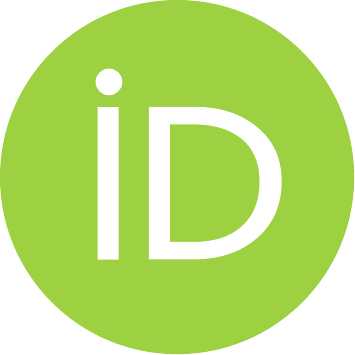}\hspace{1mm}Sayeh GHOLIPOUR PICHA}\\
	Univ. Grenoble Alpes,\\
        CNRS, Grenoble INP, GIPSA-lab,\\
        38000 Grenoble, France\\
	\texttt{sayeh.gholipour-picha@grenoble-inp.org} \\
	\And
	\href{https://orcid.org/0000-0002-6258-6970}{\includegraphics[scale=0.06]{orcid.pdf}\hspace{1mm}Dawood AL CHANTI} \\
	Univ. Grenoble Alpes,\\
        CNRS, Grenoble INP, GIPSA-lab,\\
        38000 Grenoble, France\\
	\texttt{dawood.al-chanti@grenoble-inp.fr} \\
        \And
        \href{https://orcid.org/0000-0002-5937-4627}{\includegraphics[scale=0.06]{orcid.pdf}\hspace{1mm}Alice CAPLIER} \\
	Univ. Grenoble Alpes,\\
        CNRS, Grenoble INP, GIPSA-lab,\\
        38000 Grenoble, France\\
	\texttt{alice.caplier@grenoble-inp.fr} \\
}
\renewcommand{\headeright}{}

\hypersetup{
pdftitle={How far generated data can impact Neural Networks performance?},
pdfsubject={cs.CV},
pdfauthor={Sayeh GHOLIPOUR PICHA, Dawood AL CHANTI, Alice CAPLIER},
pdfkeywords={Facial Expression Recognition, Generative Adversarial Networks, Synthetic data.},
}

\begin{document}
\maketitle
\begin{abstract}
	The success of deep learning models depends on the size and quality of the dataset to solve certain tasks. Here, we explore how far generated data can aid real data in improving the performance of Neural Networks. In this work, we consider facial expression recognition since it requires challenging local data  generation at the level of local regions such as  mouth, eyebrows, etc, rather than simple augmentation. Generative Adversarial Networks (GANs) provide an alternative method for generating such local deformations but they need further validation. To answer our question, we consider  noncomplex  Convolutional Neural Networks (CNNs) based classifiers for recognizing Ekman emotions. For the data generation process, we consider generating facial expressions (FEs)  by relying on two GANs. The first generates a random identity while the second imposes facial deformations on top of it.  We consider training the CNN classifier using FEs from: real-faces,  GANs-generated, and finally using a combination of real and GAN-generated faces. We determine an upper bound regarding the data generation quantity to be mixed with the real one which contributes the most to enhancing FER accuracy.  In our experiments, we find out that 5-times more synthetic data to the real FEs dataset increases accuracy by 16\%.
\end{abstract}


\section{\uppercase{Introduction}}
\label{sec:introduction}

Deep learning (DL) has achieved high accuracy performance in various complex tasks including recognition \cite{dl-recognition}, detection \cite{dl-detection}, localization \cite{dl-localization}, etc. Yet despite its success, it requires large amounts of labeled data, especially if high performance is required. For instance, considering a Facial Expression Recognition (FER) model trained on a specific Facial Expressions (FEs) dataset, it would not perform as well when applied to a moderately different real-world dataset. This is due to the distribution shift coming from a lack of diversity and biases in the datasets against certain demographic changes \cite{ref1} such as race, gender, and age.

Biases in the training data prone trained models towards overfitting as they are optimized over the majority samples (e.g. certain age) represented in the dataset. Hence a low performance is expected  over minor samples (e.g. certain races). To address this issue, we argue that having at disposal a diverse dataset would help in overcoming such biases and building a generalizable model. However, acquiring and labeling image and video data is a very expensive and time-consuming task and sometimes it is not even feasible. In this paper, we study the impact of synthetic data generation on the performance of neural networks. We propose to alleviate the bias issue by testing a data augmentation procedure able to generate balanced and diverse data samples.

Several works \cite{FER2}, \cite{FER18}, \cite{FER20}, and \cite{FER22} routinely performed standard data augmentation using affine transformation (e.g., translation, scaling, rotation, reflection, shearing, cropping, etc.). Standard augmentation does not bring any new information to enrich the training dataset to solve the bias problem. On the contrary, Generative adversarial networks (GANs) \cite{GoodfellawGAN} offer the opportunity, to increase the amount of training samples, and to enrich the diversity of the final training set under certain experimental data generation process. In this paper, we consider an FER task and we address and evaluate the use of generated synthetic FEs via GANs to compensate the lack of diversity in FE training databases in an attempt to reduce the bias of the considered FER model and to increase its  generalization ability.

Here, we consider a classical CNN classification scheme as it is not our intention to build a novel classifier. However we carefully design the data augmentation scheme based on combining multiple GANs that consider generating: i) new and diverse FEs with new identities and races, various genders, and different ages; ii) various FEs deformation intensities, which makes the generated facial expressions closer to spontaneous human behavior; and iii) balanced dataset where we guarantee that each identity gets the same amount of generated images per emotion class.

To this end, our contributions are:
\begin{itemize}
        \item We design a method to generate diverse and balanced facial expression deformations.
        
        \item We empirically investigate the contribution of synthetic data and their role in improving DL performance.

    \item We perform a cross-database evaluation to estimate fairly the impact of generated data on the generalizability of the trained model.
\end{itemize}

The paper is structured as follows: Section \ref{sec:review} discusses related works; Section \ref{sec:method} presents the proposed procedure of building an FER system based on augmented data; Section \ref{sec:result} discusses the experimental results; Finally, Section \ref{sec:conclusion} concludes the paper.

\section{\uppercase{Related Works}}\label{sec:review}
In most traditional research in facial expression recognition, the combination of face appearance descriptors used to represent facial expressions with deep learning techniques is considered to overcome the difficult factors for FER.
Regardless, due to the small size of public image-labeled databases, Data Augmentation (DA) techniques are often used to increase the size of the database. 
In addition to DA geometric transformations, more complex guided augmentation methods can be used for DA, such as GAN. 
In \cite{FER10}, a conditional GAN is used to generate images to augment the FER2013 dataset. A CNN is used to train the predictive model, and the average accuracy increased by 5\% after applying the GAN DA technique.
\cite{FER25} proposed an FER method based on Contextual GAN. Chu's model uses a contextual loss function to enhance the facial expression image and a reconstruction loss function to retain the subject's identity information in the expression image. Experimental results with the extended CK+ database \cite{ck} show that Chu's method improves recognition performance by 7\%. However, neither Yi's nor Chu's studies perform cross-database evaluation nor consider the generation of balanced synthetic FEs classes.
\cite{DAGAN} experimented with the combination of various data augmentation approaches, such as using synthetic images, and discovered that a combination of synthetic data with horizontal reflection, and translation can increase the accuracy by approximately 30\%. They performed cross-database evaluations by training their model on an ``augmented" KDEF database \cite{KDEF} and testing it on two different databases (CK+ and ExpW \cite{expw}).
Unlike them, we design our method to consider a diverse but balanced generation of FE classes and create our experimental setup to resemble fair performance metrics.

\section{\uppercase{Data Modality}}\label{sec:method}

Generative Adversarial Networks are used to generate different FEs for training our FER algorithm. Our model design splits into three different compartments: the data generation stage, the CNN classifier training stage, and the inference stage.

\subsection{Dataset Generation Process}\label{subsec:data_generation}

\begin{figure}[h!]\captionsetup[subfigure]{font=tiny}
\footnotesize
     \centering
     \begin{subfigure}{0.2\textwidth}
         \centering
         \includegraphics[width=0.5\textwidth]{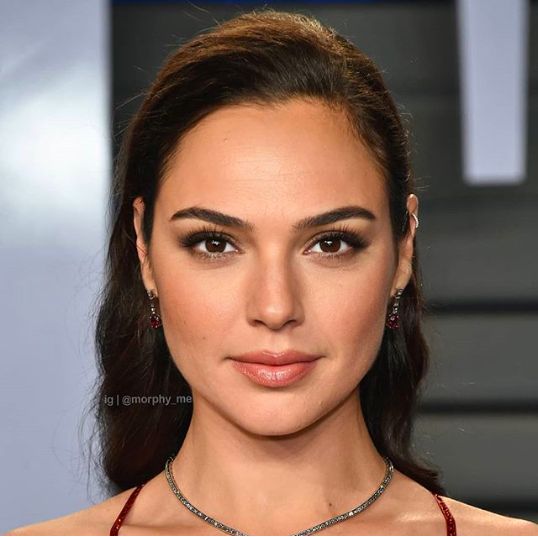}
         \caption{Real human facial features.}
         \label{fig:galgadut}
     \end{subfigure}%
     \begin{subfigure}{0.2\textwidth}
         \centering
         \includegraphics[width=0.5\textwidth]{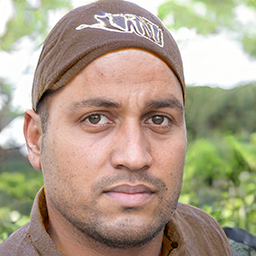}
         \caption{StyleGAN Generated identity.}
         \label{fig:generated_identitiy}
     \end{subfigure}
        \caption{Two samples of facial features.}
        \label{fig:samples}
\end{figure}

\begin{figure*}[!ht]
    \centering
    \includegraphics[trim={17.5cm 0 0 0},clip,width=0.6\textwidth]{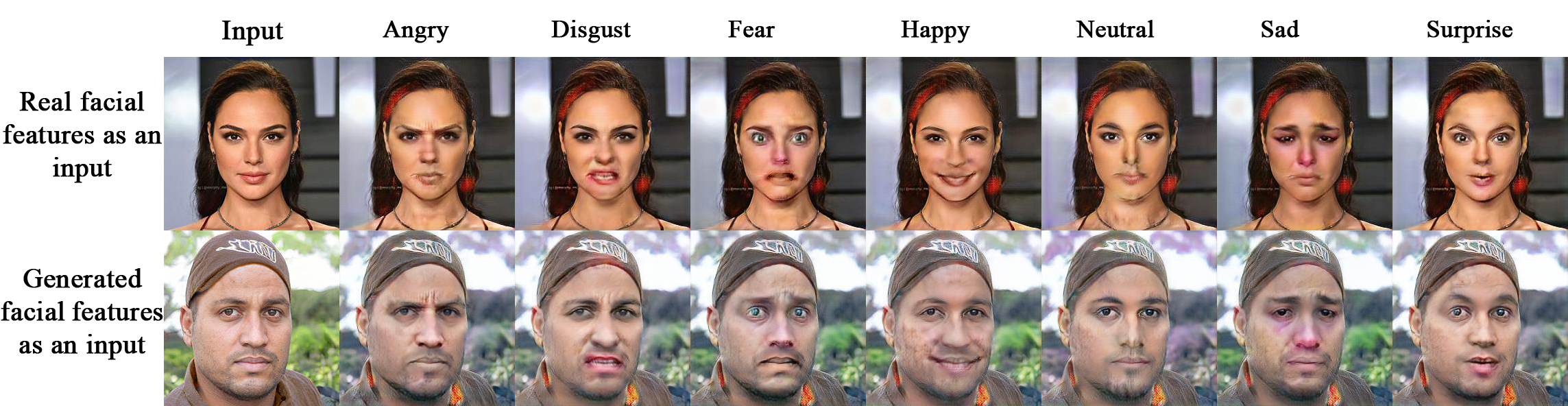}
    \caption{Examples of synthetic facial expressions. In the top row, the StarGAN model acts on an actual human face. In the bottom row, the model works with a generated identity.}
    \label{fig:stargan_output}
\end{figure*}

\begin{figure}
    \centering
    \begin{subfigure}[b]{0.5\textwidth}
         \centering
         \includegraphics[width=0.8\textwidth]{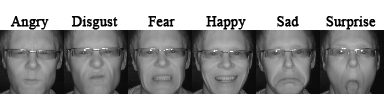}
         \caption{samples from the Oulu-CASIA database.}
         \label{fig:final_oulu}
     \end{subfigure}
     \hfill
     \begin{subfigure}[b]{0.5\textwidth}
         \centering
         \includegraphics[width=0.8\textwidth]{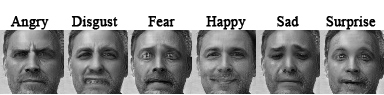}
         \caption{Samples from the generated dataset.}
         \label{fig:final_generated_identitiy}
     \end{subfigure}
    \caption{Samples of generated FEs either on a real face or on a generated one.}
    \label{fig:final_dataset}
\end{figure}

Our data generation process relies on using two GANs on top of each other. One is for new identity generation while the other is used to impose the generation of local FEs.  First, we generate new identities with new facial features using the StyleGAN model of \cite{StyleGAN2} that randomly generates realistic human faces. Additionally, since we want to compare the performance of our FER model trained with both real or generated facial features, we build a database that resembles existing public databases. In those public datasets, subjects pose different expressions in front of a fixed-setting camera.
For this reason, we build a novel method that jointly uses StyleGAN and StarGAN on top of each other as a way to reinforce the FEs generation process over new identities.
However, due to the randomness of the StyleGAN model and the desirability of a balanced training set, we use the structure of a StarGAN model \cite{stargan} for image-to-image translation with different settings to artificially synthesize the six Ekman emotions (anger, disgust, fear, happiness, sadness, and surprised) on a single generated identity.
We train the StarGAN model with the spontaneous public database Affectnet-HQ \cite{affecnet} since this database captures images from various settings, and from lots of people through the internet. We use the trained model to generate facial expressions on both real face images and StyleGAN generated face images as shown in figure \ref{fig:samples}. The final result of using the image-to-image translation StarGAN model to synthesize different expressions for a given real or generated identity is shown in figure \ref{fig:stargan_output}. As we can see, there are several artifacts in the output images, which are mainly found on the outer part of the face. However, these artifacts are not important in our task since we only focus on facial features for facial expression recognition.
During this process, we generated 100,000 identities and synthesized 6 basic emotions on each of them. Finally, with some preprocessing (face cropping, gray-scale, and resizing), we generated the balanced dataset illustrated in figure \ref{fig:final_dataset}.


\subsection{Convolutional Neural Network}\label{subsec:CNN}

In the second stage of our method, we design a CNN classifier whose architecture is summarized in table \ref{tab:cnn_classifier}. Our purpose is to use a simple yet effective classifier in order to focus our attention on the contribution of GAN-generated images with respect to model enhancement. To avoid the overfitting effect, we use drop-out layers.

In the first experiment, the CNN classifier is trained on the two facial expression public databases (RaFD \cite{RaFD}, and Oulu-CASIA \cite{oulu}) labeled with the 6 basic Ekman emotions. In the second experiment, the CNN model is trained again from scratch using only generated facial expressions. These two control experiments serve as baselines. Finally, we re-train the CNN model again by gradually augmenting the public databases with images of generated FEs with the same number of identities each time.
\begin{table}[]
    \centering
    \resizebox{0.7\textwidth}{!}{
    \begin{tabular}{|c|ccccc|}
    \hline
         & Layer & input & number of filters & Pool size  & Activation function\\
         \hline
         1 & Conv2D & $64 \times 64 \times 1$ & 32 & & Relu\\
         2 & Conv2D & $64 \times 64 \times 32$ & 64 & & Relu\\
         3 & Max Pooling & $64 \times 64 \times 64$ & & $2 \times 2$ & \\
         \hline
         4 & \multicolumn{4}{c}{Drop out 25\%}&\\
         \hline
         5 & Conv2D & $32 \times 32 \times 64$ & 128 & & Relu\\
         6 & Max Pooling & $32 \times 32 \times 128$ & & $2 \times 2$ & \\
         7 & Conv2D & $16 \times 16 \times 128$ & 128 & & Relu\\
         8 & Max Pooling & $16 \times 16 \times 128$ & & $2 \times 2$ & \\
         \hline
         9 & \multicolumn{4}{c}{Drop out 25\%}&\\
         \hline
         10 & \multicolumn{4}{c}{Flatten}&\\
         11 & \multicolumn{4}{c}{Dense \textbf{1024}} & Relu\\
         \hline
         12 & \multicolumn{4}{c}{Drop out 50\%}&\\
         \hline
         13 & \multicolumn{4}{c}{Dense \textbf{6}} & Softmax\\
         \hline
    \end{tabular}}
    \caption{Model summary of the considered CNN-based model for Facial Expression Recognition.}
    \label{tab:cnn_classifier}
\end{table}

\subsection{Testing Phase}\label{subsec:test_phase}

To fairly evaluate the performance of our method, we split the real and generated database to create a test database. About 17\% of the data that are created are used for testing purposes. Although these test datasets are necessary to assess the model's performance, the similarity in distribution between the test and train sets makes it difficult to determine the exact contribution of the generated data to the model's performance. To ensure a fair analysis of the results and prevent bias in model prediction, it is important for our settings to perform a cross-database evaluation in which our test data resemble zero correlation with training sets. Hence, to have a fixed reference test dataset to compare all the models, we use the MMI database \cite{MMI} which is completely blind to the training process.

\section{\uppercase{Results and Analysis}}\label{sec:result}

In the following, we present the results of three experiments with different settings and designs. RaFD \cite{RaFD} and Oulu-CASIA \cite{oulu} real databases have been used in experiments 1 and 3. Alternatively, in experiment 2 we use only synthetic data, and these synthetic data are also used in experiment 3 for further analysis.

\subsection{Experiment 1 - Training with real data}\label{subsec:exp1}
In this first baseline experiment, the CNN classifier for the FER model has been trained with real face images coming from RaFD and Oulu-CASIA which in total have 144 subjects. Here, we use 109 subject images with 6 basic emotions for training, 10 for validation, and 25 for testing. From these data, we only consider the frontal faces that are associated with emotional labels.

\begin{figure}
    \centering
    \includegraphics[width=6.5cm]{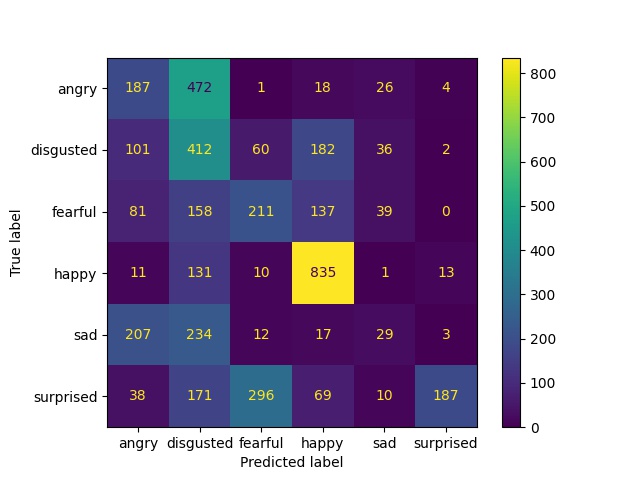}
    \caption{Confusion Matrix on the MMI database in experiment one.}
    \label{fig:MMI_CM_real}
\end{figure}

\begin{figure}
    \centering
    \includegraphics[width=6.5cm]{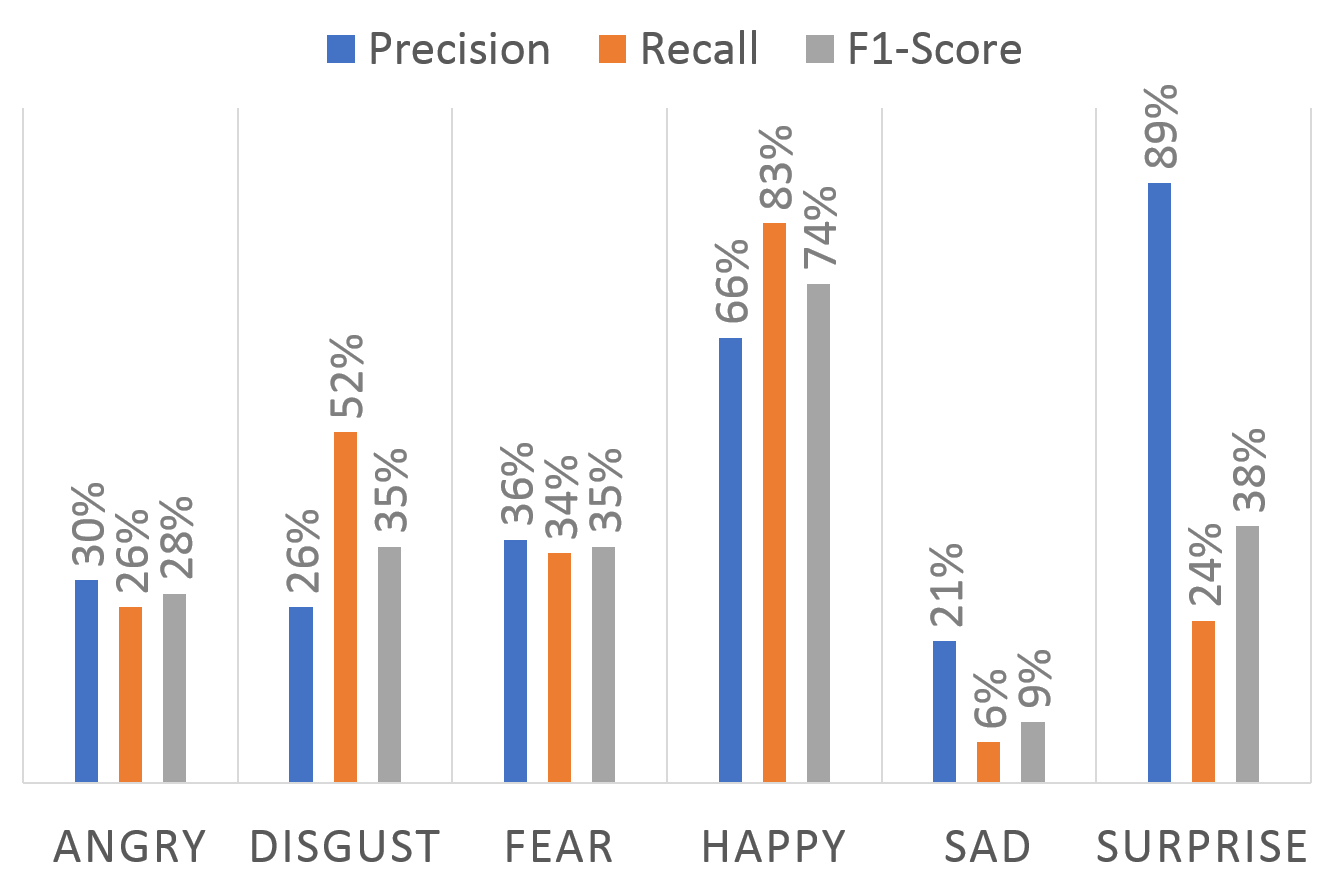}
    \caption{Precision, recall, and F1-score on the MMI database  obtained on the CNN model trained with real faces only (cf. experiment 1).}
    \label{fig:real_mmi_score}
\end{figure}

Our trained CNN classifier achieves 69.6\% of accuracy when it is tested over 25 subjects. However, we observe overfitting because our model achieves 84.5\% in the training phase.
By applying a cross-database evaluation using the MMI database \cite{MMI}, the obtained accuracy drops to 45\% which is expected due to the limited number of subjects we have in the training dataset. Next, we analyze each class separately to get an insight into the classes' separability. 
Figure \ref{fig:MMI_CM_real} presents the associated confusion matrix.
Based on this map, the ``Happy" emotion has the best performance and the ``Disgust" class is also showing a good performance compared to other classes of emotions. While this confusion matrix provides a comprehensive overview of each class, it is unnormalized, making later comparisons difficult. 
To analyze the results of the cross-database evaluation on the MMI database in a more detailed manner, in a second step, we measure three metrics (precision, recall, and F1-score) to explore the model's prediction with the annotations provided for the MMI database.
Figure \ref{fig:real_mmi_score} presents these metrics results for each class of emotions. It can be noticed that the model trained on real facial features has the most difficulty at recognizing the ``Sad" class, while the other classes are not showing a good performance either. Despite having two real databases for training, this model fails to perform adequately. As a result, adding more data to the training is necessary and we argue that adding synthetic data might aid in overcoming the overfitting and also in lifting up the accuracy rate. 

\subsection{Experiment 2 - Training with synthetic generated data}\label{subsec:exp2}

In the second experiment, we follow the same protocol. The only difference is that we are using synthetic facial images as the training dataset. We consider the same number of synthetic identities as in the real dataset in the first experiment (109 identities). These identities are generated with the process presented in section \ref{subsec:data_generation}. For each identity, all six basic emotions exist in the dataset.
Training our model with this synthetic dataset, the accuracy reaches 99.84\% during the training process and 97.6\% while testing on the synthetic dataset. Also, no overfitting is observed in this experiment. 
Although these results show a significant improvement, performing the cross-database evaluation on the MMI database is not that promising. On the MMI database, the obtained accuracy drops to 47\% showing nearly the same performance as the model trained in experiment 1.

\begin{figure}
    \centering
    \includegraphics[width=6.5cm]{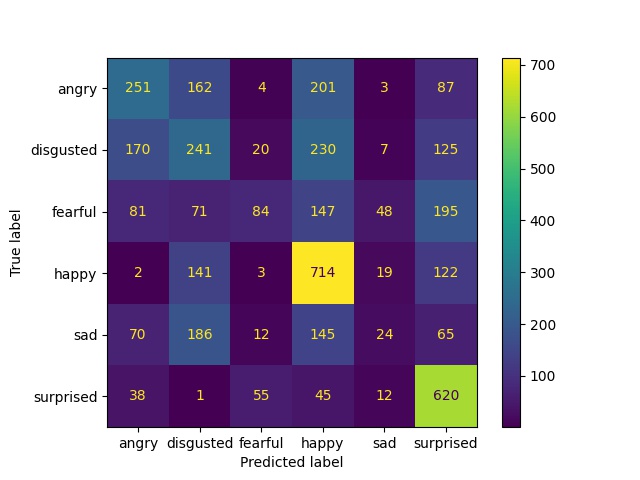}
    \caption{Confusion Matrix on MMI database in experiment two.}
    \label{fig:MMI_CM_GAN}
\end{figure}

\begin{figure}
    \centering
    \includegraphics[width=6.5cm]{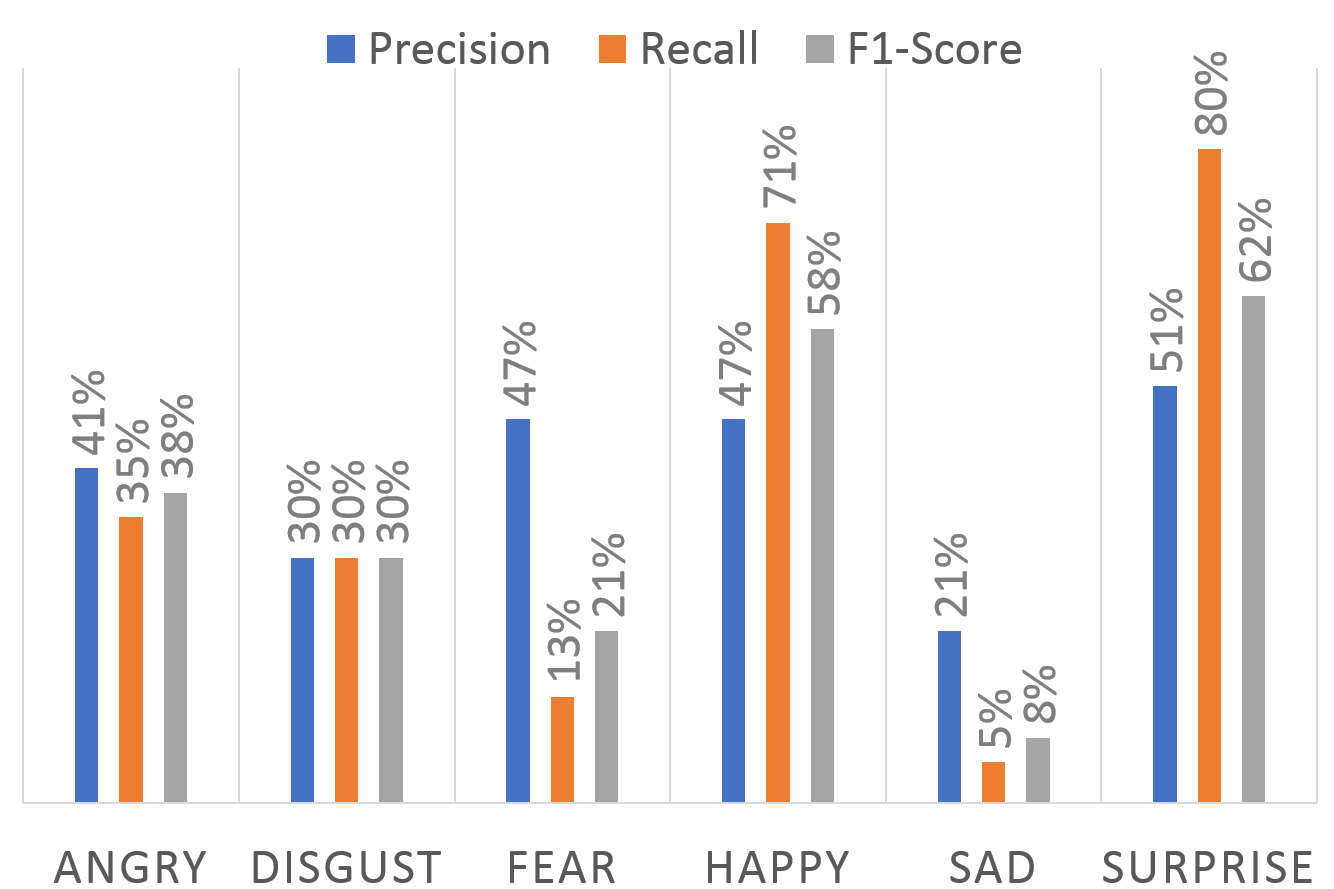}
    \caption{Precision, recall, and the F1-score on the MMI database for the model trained in  experiment two (section \ref{subsec:exp2}).}
    \label{fig:gan_mmi_score}
\end{figure}

Figure \ref{fig:MMI_CM_GAN} presents the confusion matrix of the model trained with the synthetic dataset for the MMI database to discover whether the model trained on synthetic data has similar classifying difficulties as the model trained on real data. 
Comparing the Confusion Matrix in figures \ref{fig:MMI_CM_real} and \ref{fig:MMI_CM_GAN} we notice:
\begin{enumerate}
    \item There has been a huge improvement in recognizing the class ``Surprised", 620 samples instead of 187 samples.
    \item We can observe improvements in the recognition of the ``Angry" class, 251 samples instead of 187.
    \item As compared to the CNN classifier trained on real faces, we see some drop in the ``Disgust", ``Fear", and ``Sad" classes, but both models seem to have similar difficulties.
\end{enumerate}
Also based on the presentation of precision, recall, and F1-score in figure \ref{fig:gan_mmi_score}, it appears the recall scores have decreased for most classes. We can therefore say that except for the ``Surprised" class, the CNN model trained on synthetic data alone is unable to match the actual facial expressions annotations provided in the MMI database. 
The results of our current experiment prove that synthetic datasets can achieve similar performance as real datasets. 
Hence our final aim is to increase the dataset size to improve the performance at all class levels. In this case, we hope that the combination of these two databases will help solve such problems.

\subsection{Experiment 3 - Training with augmented datasets}\label{subsec:exp3}
In the last experiment, we augment the Real Facial Expressions (RFEs) dataset of experiment 1 with Generated Facial Expressions (GFEs).
The number of generated identities in each unit is the same as the number in the real database used for experiment 1 (109 identities for training, 10 identities for validation, and 25 identities for testing). 
As an example, RFEs + 2 $\times$ GFEs is the extension of the real FEs with two units of generated FEs (109 real identities + 218 generated identities for training). Each of the augmented datasets is split into training, validation, and test sets. And the CNN model is trained on each dataset individually. Each augmented dataset is represented in table \ref{tab:DA_per} indicating the model accuracy during training and testing. 
The results demonstrate that adding more synthetic FEs to the training set results in constant improvement of training and testing accuracies. The study also reports no evidence of overfitting.
\begin{table}[]
    \centering
    \begin{adjustbox}{width=0.4\textwidth}
    \begin{tabular}{cccc}
    \hline
         & Training accuracy& Testing accuracy\\
         \hline
         RFEs + GFEs & 91\% & 85.3\%\\
         RFEs + 2 $\times$ GFEs& 93.8\% & 89\%\\
         RFEs + 3 $\times$ GFEs& 94.8\% & 92.7\%\\
         RFEs + 4 $\times$ GFEs& 95.8\% & 92.5\%\\
         RFEs + 5 $\times$ GFEs& 97.6\% & 94.3\%\\
         RFEs + 6 $\times$ GFEs& 97.8\% & 94\%\\
         RFEs + 10 $\times$ GFEs& 97.9\% & 95.1\%\\
         RFEs + 15 $\times$ GFEs& 98.9\% & 95.5\%\\
         RFEs + 20 $\times$ GFEs& 98.9\% & 97\%\\
         \hline
    \end{tabular}
    \end{adjustbox}
    \caption{Accuracy of the model trained on augmented datasets (Real Facial Expressions (RFEs) augmented by Generated Facial Expressions (GFEs)).}
    \label{tab:DA_per}
\end{table}

\begin{figure*}
    \centering
    \includegraphics[width=0.9\textwidth]{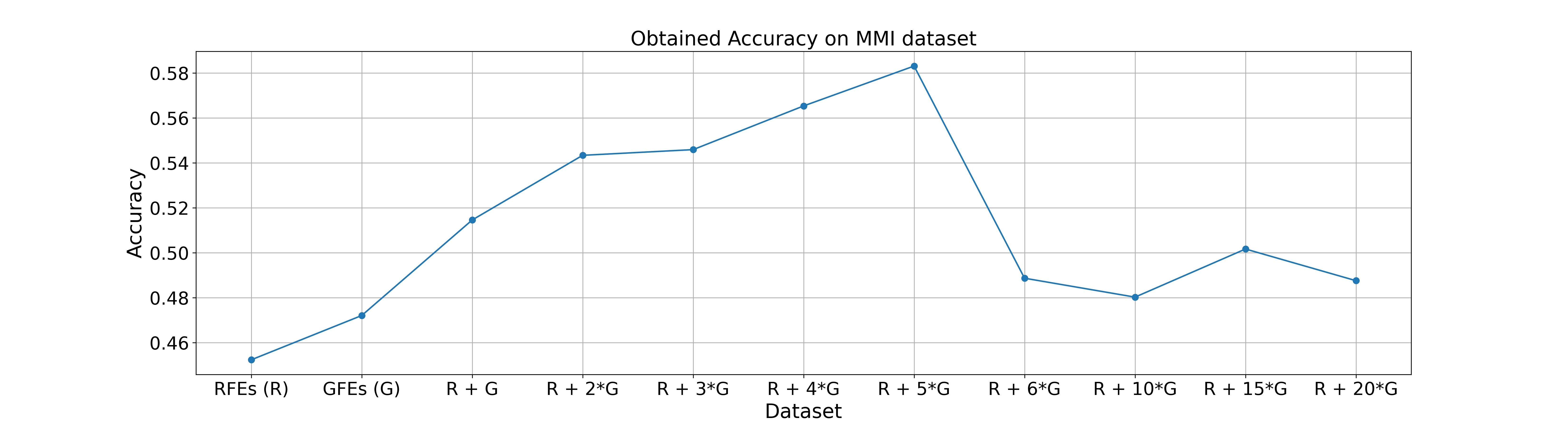}
    \caption{The result of cross-database evaluation on the MMI database. RFEs is referring to Real Facial Expressions and GFEs is referring to Generated Facial Expressions.}
    \label{fig:MMI_acc}
\end{figure*}

The cross-database evaluation on the MMI database is then performed for further validation and figure \ref{fig:MMI_acc} shows the accuracy obtained from each trained model. Note that the first two points are the result of cross-database evaluation obtained in experiments 1 and 2 respectively. According to this figure, the highest accuracy corresponds to the model trained on the RFEs + 5 $\times$ GFEs dataset with 58.3\%. This performance from the model trained on the 5th augmented dataset indicates a 13\% gain in response to the model trained in experiment 1 (with real FEs). But beyond this point, the accuracy drops significantly due to a catastrophic forgetting mode caused by the large number of synthetic facial features in the training set overwhelming the facial features of the real face database.

\begin{figure}
    \centering
    \includegraphics[width=6.5cm]{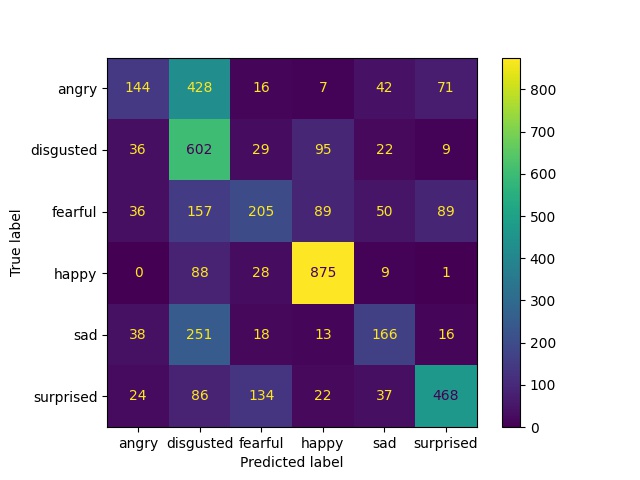}
    \caption{Confusion Matrix on MMI database in experiment three with the model trained on RFEs + 5 $\times$ GFEs database.}
    \label{fig:MMI_CM_MIX5}
\end{figure}

\begin{figure}
    \centering
    \includegraphics[width=6.5cm]{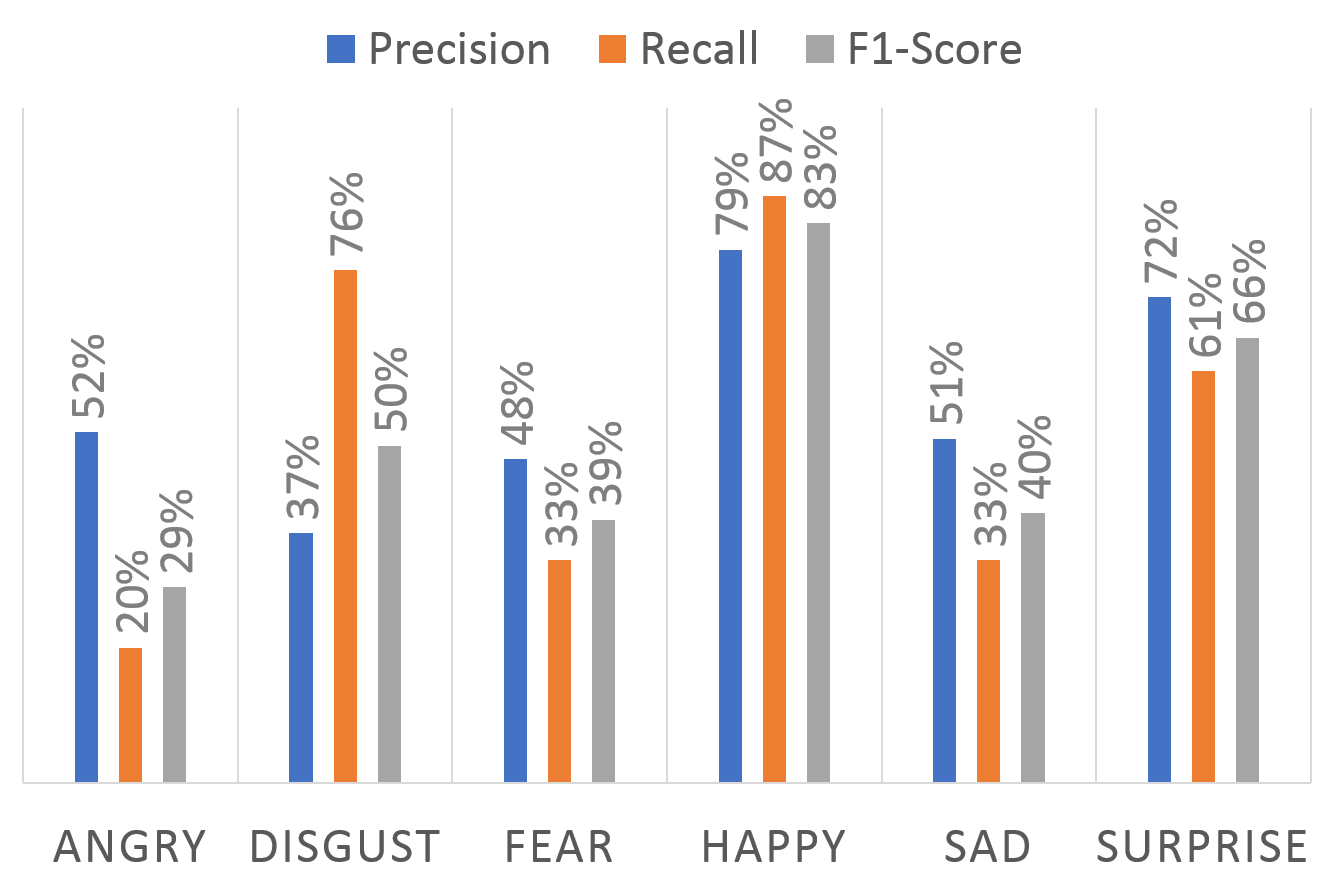}
    \caption{Precision, recall, and F1-score on the MMI database for the model trained on the RFEs + 5 $\times$ GFEs database in experiment three (section \ref{subsec:exp3}).}
    \label{fig:MIX5_mmi_score}
\end{figure}

To see the improvement of our best model (RFEs + 5 $\times$ GFEs) in each class separately, we present the confusion matrix and the calculated precision, recall, and F1-score metrics on the MMI database in figures \ref{fig:MMI_CM_MIX5} and \ref{fig:MIX5_mmi_score} respectively.
It appears that the ``Sad" class performs significantly better than the two baseline experiments in all three metrics.
Based on the recall scores in all classes, we can conclude that this trained model matches facial expressions in the MMI database to their actual annotations better than other trained models. In contrast with the model trained in experiment 1 (training set of real FEs), only the ``Anger" class's performance decreases. In conclusion, based on our observation, we can say that generated data along with the real facial features is helping the model's recognition ability.
\begin{table*}[]
    \centering
    
    \begin{adjustbox}{width=0.7\textwidth}
    \Large
    \begin{tabular}{|p{0.2cm}|p{4.5cm}|p{12.5cm}|p{2cm}|}
    \hline
         &Method&Training Database&Accuracy\\
         \hline
         1&Proposed method & RaFD + Oulu-CASIA +GAN & \textbf{87.49\%} \\
         \hline
         2&\cite{DAGAN} & KDEF & 83.30\% \\
         \hline
         3&\cite{FER2} & MUG & 45.60\% \\
         \hline
         4&\cite{FER2} & JAFFE & 48.20\% \\
         \hline
         5&\cite{FER2} & BOSPHOROUS & 57.60\% \\
         \hline
         6&\cite{FER17} & KDEF \cite{KDEF} & 78.85\% \\
         \hline
         7&\cite{FER18} & JAFFE & 54.05\% \\
         \hline
         8&\cite{FER20} & MMI+FERA & 73.91\% \\
         \hline
         9&\cite{FER21} & MultiPIE \cite{multipie}, MMI, CK+, DISFA \cite{disfa}, FERA \cite{fera}, SFEW \cite{sfew}, and FER2013 & 64.20\% \\
         \hline
         10&\cite{FER22} & CK+, JAFFE, MMI, RaFD, KDEF, BU-3DFE \cite{BU-3DFE}, and ARFace \cite{arface} & \textbf{88.58\%} \\
         \hline
    \end{tabular}
    \end{adjustbox}
    \caption{Comparison among state-of-the-art cross-database experiments tested on the CK+ database.}
    \label{tab:comp_literature}
\end{table*}

Furthermore, there is no limit to the number of identities we can generate.
But there is a point beyond which adding new generated FEs no longer improves the results. We have observed experimentally that there is an upper limit in augmentation in relation to the size of the real face database.

\subsection{Comparison with the state-of-the-art}
As a final step in this study, we compare our results with state-of-the-art findings. We use the VGG16 tool to calculate the accuracy of the FER VGG16 model on the MMI database. With that model, we achieve 54.08\% accuracy while our best CNN-based model reaches 58.3\% in accuracy. Through the use of synthetic facial features and a simpler model, we enhance the accuracy by 4\%.

Many state-of-the-art studies have reported their evaluation results on the CK+ database. Nevertheless, we did not use the CK+ database in our training or testing processes. Therefore, in order to perform the comparison, we evaluate our best model performance on the CK+ database and the result is presented in table \ref{tab:comp_literature}. 
It can be seen that the approach proposed by \cite{FER22} is the only one that outperforms our proposed CNN model. However, the difference is only 1.09\% while they trained their model using 6 different public databases and some classical data augmentation techniques. Whilst our results are achieved with smaller training datasets using only two public databases and GAN images which makes our results more outstanding. In addition, compared to the study in \cite{DAGAN} that is explained in section \ref{sec:review}, even though their model's accuracy increased by 30\%, our model had more promising results based on this cross-database evaluation.

\begin{figure}
    \centering
    \includegraphics[width=5.5cm]{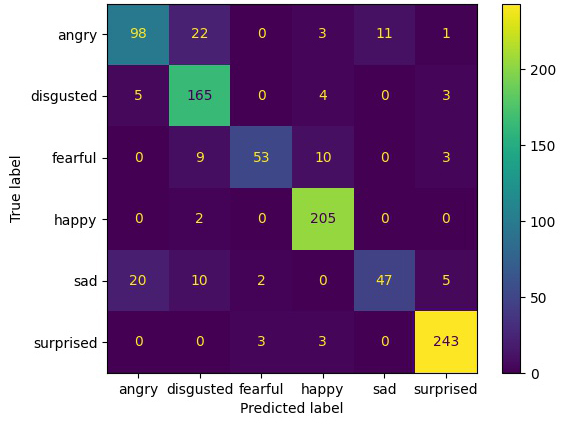}
    \caption{Confusion Matrix on CK+ database.}
    \label{fig:CK_CM}
\end{figure}

Figure \ref{fig:CK_CM} shows the result of our CNN-based model in the cross-database evaluation on the CK+ database. For this database, we achieve an accurate model for most of the classes even though there is no record of this public database in our training data.

During this study, we replaced the MMI database with the CK+ database for cross-database evaluation. As a result, model performance increased by 16\% rather than the 13\% gain we previously achieved. While it is undeniable that generated data is a costless method that can help improve FER model accuracy, the exact gain would be determined by the test databases in applications.

\section{\uppercase{Conclusions}}
\label{sec:conclusion}

The purpose of this study was to investigate how generated data could be used to augment the data in a deep learning model to improve its performance. We chose a simple facial expression recognition model for this proposal. Our synthetic balanced dataset was created using two GAN models to test the potential improvement of the FER model performance. With real databases, synthetic datasets, and augmented datasets, we trained the CNN classifier multiple times for the FER task.

Our study confirms that enriching the training dataset with GAN images can improve CNN classifier performance.
Training and cross-database evaluation performances were improved by augmenting real databases with synthetic facial features.
In comparison to a model trained solely from real facial images, our best model shows a 16\% increase in accuracy.
On the same database, we also compared our results with the state-of-the-art and computed the accuracy of the VGG16 model, achieving 4\% higher accuracy.

For further study, we propose to first augment the training database with additional real facial expressions. This will enable us to improve the performance of the model, as it would also let us augment more GAN images. Secondly, we propose to enrich the VGG16 database with our generated dataset to see if we can improve the performance of the VGG16 model as well. And third, we would like to study the potential performance increase for other applications related to facial models.
\paragraph*{\textbf{Material, codes, and Acknowledgement:}} Results can be reproduced using the code available in the GitHub repository \url{https://github.com/sayeh1994/synthesizin_facial_expression} and \url{https://github.com/sayeh1994/Facial-Expression-Recognition}. Most of the computations presented in this paper were performed using the \cite{gricad} infrastructure (\url{https://gricad.univ-grenoble-alpes.fr}), which is supported by Grenoble research communities.

\bibliographystyle{unsrtnat}
\bibliography{references}  






\end{document}